\documentclass[11pt,letterpaper]{article}
\usepackage{cogsys}
\usepackage[T1]{fontenc}
\usepackage{times}
\usepackage[pdftex]{graphicx} %

\usepackage{natbib}
\usepackage{algorithm}
\usepackage[noend]{algpseudocode}
\usepackage{subcaption}
\usepackage[flushleft]{threeparttable}

\ShortHeadings{Toward Constraint Compliant Goal Formulation and Planning}
              {S.\ Jones and R.\ Wray}

\begin{document} 

\title{Toward Constraint Compliant Goal Formulation and Planning}

\author{Steven J. Jones}{steven.jones@cic.iqmri.org}
\author{Robert E. Wray}{robert.wray@cic.iqmri.org}
\address{Center for Integrated Cognition, IQM Research Institute, 
         Ann Arbor, MI 48105 USA}
\vskip 0.2in

\begin{abstract}
One part of complying with norms, rules, and preferences is incorporating constraints (such as knowledge of ethics) into one's goal formulation and planning processing. We explore in a simple domain how the encoding of knowledge in different ethical frameworks influences an agent's goal formulation and planning processing and demonstrate ability of an agent to satisfy and satisfice when its collection of relevant constraints includes a mix of ``hard'' and ``soft'' constraints of various types. How the agent attempts to comply with ethical constraints depends on the ethical framing and we investigate tradeoffs between deontological framing and  utilitarian framing for complying with an ethical norm. Representative scenarios highlight how performing the same task with different framings of the same norm leads to different behaviors. Our explorations suggest an important role for metacognitive judgments in resolving ethical conflicts during goal formulation and planning. 
\end{abstract}

\section{Introduction}

Artificial agents exist in environments defined by more than their tasks. Constraints on behavior arise from many sources, ranging from physical limitations to social norms. Importantly, in real-world, human-populated environments, constraints may arise that are not directly relevant to (or readily anticipated from) the task domain alone. Instead, the confluence of domain, task, and the overall human context (including laws, norms, etc.) introduce systems of constraints that will often include interactions and inconsistencies. For example, it might not be possible for an agent to satisfy all relevant constraints and also perform a task as prescribed.
Our current research focuses on identifying, researching, and meeting the myriad computational challenges that must be met to enable AI agents to achieve their tasks while conforming to such constraints \citep{wray_computational-level_2023}. 

When a human generates and evaluates a course of action, they (usually) take into account the ethical norms pertinent to their situation. Or, perhaps more importantly, we can say that there is a societal expectation that they will do so  \citep[and punishment is apt for those who do not,][]{sripada_framework_2006,molho_direct_2020}. 
Further, humans can generate normatively-acceptable behavior even in the presence of many interacting and conflicting constraints \citep{sripada_framework_2006,aldinhas_ferreira_networks_2017}. In a world populated by various artificial agents -- robots, virtual assistants, work teammates, etc. -- humans will expect these artificial agents to conform to norms, similar to how they expect other humans to behave. Thus, it is critical that artificial agents have the capacity to generate plans and behavior that meet these constraints \citep{arkin_moral_2011,rossi_building_2019,langley_explainable_2019,wray_incorporating_2021,giancola_ethical_2020,misselhorn_artificial_2020,malle_requirements_2019}.

As part of this larger effort to investigate and understand the means by which artificial agents can achieve constraint compliance, this paper focuses on the capability of an agent to plan its tasks while taking into account these constraints and how they impinge upon an agent's tasks. We assume that the world is sufficiently uncertain (many constraints can arise from many sources) and unstable (conceptual drift) that \textit{a priori} training is not a (complete) solution. Instead, in our conception of the problem, current constraints influence both the formulation of goals (what the agent should accomplish) and planning (how the agent should act to achieve its goals). We emphasize how an agent can plan in the presence of ethical norms, especially those that impose ``soft constraints'' on resulting behavior. Examples of the class of norms we consider include interpersonal courtesy (respecting personal space) and behavior mirroring (e.g., being deliberately quiet in a library; matching speed of others when driving).

In the rest of the paper, we present the design for and implementation of an agent that can  ``take into account'' norms (and other constraints) when planning. We characterize this problem in more detail by identifying computational requirements and introducing an example domain in which to compare and contrast alternative approaches to reasoning about ethical norms. We then summarize the implementation and some of the implications of different design decisions and different ethical knowledge on agent behavior. Importantly, we illustrate how different assumptions about how norms are represented and interpreted (the ``ethical framework'') influence both the resulting behavior and computational demands on the agent. Further, when faced with dilemmas where tasks and norms conflict, metacognition is needed for goal formulation and planning. Furthermore,  the specific content of knowledge needed to support such metacognition is contingent on the ethical frame used in planning.

\section{Background} %
\begin{figure}[tb]
    \centering
    \includegraphics[width=\textwidth]{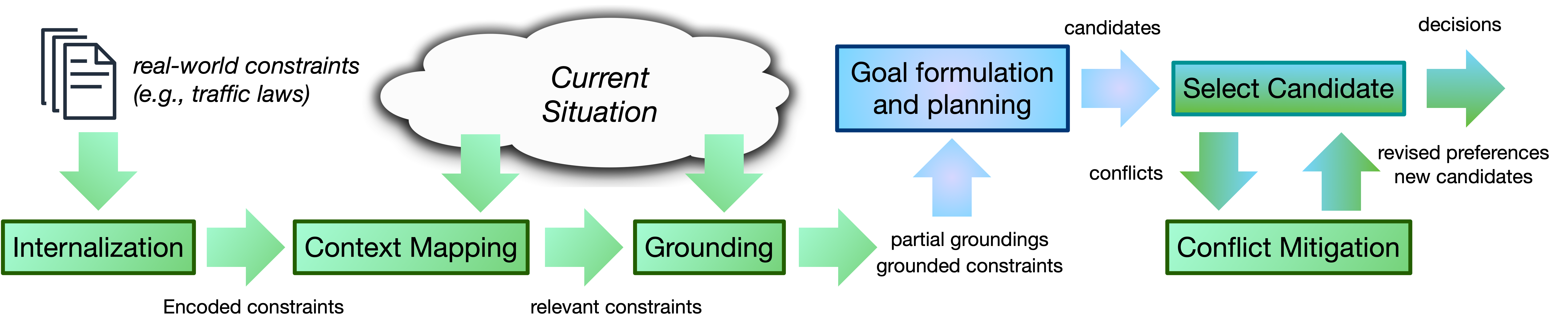}
    \caption{A computational pipeline for constraint compliance \citep[adapted from][]{wray_computational-level_2023}.}
    \label{fig:constraint-compliance}
\end{figure}

Wray et al \citeyearpar{wray_computational-level_2023}  proposed and described an implemented computational pipeline for complying with constraints, illustrated in Figure~\ref{fig:constraint-compliance}. This end-to-end pipeline is  implemented in Soar \citep{laird_soar_2012}.  ``Placeholder'' solutions have been developed for each of the core computational functions (boxes in the figure). For instance, for \textit{Internalization}, we have developed a language and tool that supports the declarative assertion of constraints, which are then encoded in Soar's semantic memory.

The pipeline is designed to enable further exploration of more sophisticated approaches to individual computational functions. For example, in recent work, we developed a general algorithm for grounding abstract constraint specifications to an agent's situation \citep{jones_challenge_2023}, which is necessary for evaluating potentially unsatisfied constraints (e.g., which might arise when a constraint can only be partially grounded in the current situation). The important observation for this paper is that this Constraint Compliance pipeline provides a tool for exploring end-to-end agent behavior, including reasoning about constraints and performing tasks in a task domain, while enabling research focus on individual competencies within the pipeline.

\begin{table*}[t]
    \centering
    \caption{Requirements for goal formulation and planning with systems of constraints. }
    \begin{small}
    \begin{threeparttable}
    \begin{tabular}{p{1.05in}p{4.6in}}
    Requirement & Brief Description \\ \hline
Integrate constraints from different classes of sources & Planning must take into account constraints from different sources and/or constraints that reflect different kinds of limitations. A physical constraint cannot be overcome (e.g., maximum speed), while a rule or norm (safe speed, speed limit) could be. \\ \hline
Accommodate both soft and hard constraints & Humans handle a variety of constraints, including strict prohibitions, duties, and physical barriers (``hard constraints''). Humans can also 
    balance various resource limitations, ``gray areas,'' and recommendations or preferences rather than specific prohibitions. We can refer to these as ``soft constraints''\tnote{*} \citep{rossi_soft_2006}.  Goal formulation and planning must consider the impact and influence of ``soft'' as well as hard constraints. \\ \hline
Anticipate \par interactions and failures. & An agent cannot assume that the constraints relevant to its situation are consistent or coherent or will not conflict with its task(s). In many situations, planning cannot satisfy all constraints and the task. The planning algorithm must have mechanisms that anticipate and respond to failures.  \\ 
    
    \hline \hline
         Responsive, Anytime  &  An  agent's algorithms must be responsive to environmental changes.   \\ \hline
         Accommodate partial \par observability & The agent cannot assume that its sensing is sufficient to observe all variables or states relative to its planning. \\ \hline
         Recognition Learning & The agent's planning performance should improve with experience in the domain. Generally, processing should shift from deliberative (``System 2'') to recognition (``System 1'') processing as problem solving becomes familiar \citep{kahneman_thinking_2011}. \\ \hline
         Online, \par Incremental Learning & Because the agent must be responsive to its environment (and potentially to changes in the dynamics of the environment itself; drift), the agent should learn during performance (``online'').\\ \hline
    \end{tabular}
    \begin{tablenotes}
        \item[*] \footnotesize As a technical term, ``soft constraints'' can refer to a particular formalism, but we use it colloquially.
    \end{tablenotes}
    \end{threeparttable}
    \label{tab:planning-requirements}
    \end{small}
\end{table*}

The research we report, as mentioned in the previous section, focuses on the goal formulation and planning part of the pipeline. This box has a different color in the figure because a general solution requires consideration of both general constraints and the specific demands of current tasks. Thus, our research goals are to identify how (algorithmically) constraints should influence goal formulation and planning. Obviously, planning algorithms already take constraints into account for planning. We see three key requirements, which are highlighted by the challenge of realizing planning that takes ethical constraints into account. These requirements are summarized in the first three rows of Table~\ref{tab:planning-requirements}.
While we emphasize these three requirements in the remainder of this paper, other requirements influence the designs and implementations that follow. The remainder of Table~\ref{tab:planning-requirements} lists additional requirements and briefly outlines them. Many of these requirements derive from general requirements for Constraint Compliance and are discussed further in Wray et al. \citeyearpar{wray_computational-level_2023}{}.

\section{Experimental Domain and Task}
To illustrate these requirements and the challenge of goal formulation and planning with constraints more concretely, this section introduces a domain, tasks, and examples of potential constraints that may arise in this domain.

The example domain is a maritime warehouse. The warehouse stores various cargoes of different types. Within an actual maritime warehouse, there are physical constraints (e.g., the size, weight, properties, and locations of various cargoes), constraints from law and formal procedures \citep[e.g.,][]{uscode_navigation_2024}, and norms that influence interpersonal interactions. For example, in a warehouse populated by both people and robots,  we assume that robots should generally conform to norms associated with physical movements around others. In a large warehouse, it would be unusual to move very close to another person unless you plan to engage them in some way (check their ID, ask about the status of some cargo, etc.). 

We envision a robotic agent that can perform many different tasks within this domain. For example, the robot could perform various security functions (e.g., conducting patrols, checking personnel identification, escorting personnel, etc.). For this paper, we focus on a single task: patrol. The robot should periodically circumnavigate the perimeter of the room. Such patrols are routine and required in some maritime warehouse settings \citep{uscode_navigation_2024}.

A simulation of this domain is represented in a later description of our approach in Figure~\ref{fig:utilitarianism}. Similar to other researchers \citep{loreggia_making_2022}, we adopt a relatively simple grid representation of a warehouse environment. While  a gross simplification of the complexity in an actual warehouse, the environment includes both objects and people and can support many tasks and multiple types of constraints.

For the patrol task, ideally, the robot visits every cell along the perimeter as quickly as it can move. However, we define the task with two minimum performance criteria: 1) the agent should visit some specified fraction of the total perimeter cells, and 2) the agent must complete its patrol in no more than some prescribed period of time. (As a further simplification for planning, we use the number of movements as a proxy for time: moving to an open cell consumes one unit of time.)

Objects in the environment impose different types of constraints. Crates (brown cubes) present a physical constraint: the robot cannot move into these cells. Barrels (blue/orange icons) present a physical constraint that allows movement but at a slower speed. For this paper, we assume traversing a cell with a barrel takes twice as long as traversing an open cell. 

Alongside physical constraints, we introduce the following norm: an agent should avoid entering a cell occupied by a person. This norm might be representative of courtesy or respect of the space of others. Given this norm, the result of planning is generally a selection either to move through a cell with a person to be more efficient at patrol or to go around people. Much of the remainder of the paper discusses how this constraint could be represented within a planning algorithm.%

In the examples that follow, we compare and contrast behavior in two classes of scenarios, which we term ``consensus'' and ``dilemma.'' Consensus scenarios are meant to present an obvious or easy situation in which there is no ambiguity about the appropriate path the robot should take, regardless of how it interprets the norm. Figure~\ref{fig:utilitarianism}(a) is an example. In contrast, dilemma scenarios (as in Figure~\ref{fig:utilitarianism}(b)) are designed to make alternative solutions apparent. The one preferred by the robot will depend on assumptions that the robot (or the robot's designers) make.

\section{Planning with Ethical Constraints}
In this section, we describe how an agent can plan its task while incorporating various kinds of constraints. As we will see, norms require some framework for interpreting and assessing them, and the choice of framework can qualitatively change how the agent behaves with the same norm(s). We introduce a simple planning approach and then discuss how we can elaborate it to realize different types of ethical frameworks for incorporating norms.

\subsection{Baseline Planning Algorithm: A*}

We use A* search to implement navigation planning. We chose A* for three reasons: 1) it is familiar and focuses the research (and presentation) on how to incorporate norms (rather than the development of a sophisticated planning solution for this task), 2) we plan to contrast different approaches, meaning that optimizing absolute performance is not required, and 3) Soar provides an existing A* algorithm in its ``default knowledge,'' speeding implementation (we could use A* more or less immediately).

Algorithm~\ref{alg:pseudocode} summarizes the A* implementation. The boxed text represents those parts of A* that we modified to accommodate norms (discussed in the next section).  Generally, A* iteratively evaluates search nodes, computing for each node an estimated total cost $f$. $f$ is the sum of the cost to that node, $g$, and an admissible heuristic estimate of the cost to the goal from that node, $h$, where $f=g+h$. 

\begin{algorithm}[tbh]
\begin{algorithmic} [1]
\State \textbf{aStar}(start,goal)
\State openSet = priorityQueueSet($\{\}$); cameFrom = map($\{\}$); g = map($\{\}$); f = map($\{\}$);
\State g[start] = 0
\State f[start] = h(start)
\State openSet.add(start,f[start])
  \While{openSet \textbf{is not empty}}
    \State current = openSet.top()
    \If{current \textbf{is} goal}
      \State\Return reconstructPath(cameFrom,current)
    \EndIf
    \State openSet.pop()
    \For{adjacentCell \textbf{in} current.options()}
      \If{movement to adjacentCell is \fbox{legal}} \hfill \# Deontology changes ``legality''
        \State tentativeG = g[current] + \fbox{cost(current,adjacentCell)} \hfill \# Util. changes ``cost'' 
        \If{(adjacentCell \textbf{not in} g) \textbf{or} (tentativeG < g[adjacentCell])}
          \State cameFrom[adjacentCell] = current
          \State g[adjacentCell] = tentativeG
          \State f[adjacentCell] = tentativeG + h[adjacentCell]
          \If{adjacentCell \textbf{not in} openSet}
            \State openSet.push(adjacentCell,f[adjacentCell])
          \EndIf
        \EndIf
      \EndIf
    \EndFor
  \EndWhile
\State\Return failure
\end{algorithmic}
\caption{Pseudocode for A*, adapted from Wikipedia \citeyearpar{enwiki:1214713509}{}.}
\label{alg:pseudocode}
\end{algorithm}

To adapt A* for the patrol task (where the start and end locations are the same), we developed a custom heuristic rather than using  Manhattan distance, which is often used for grid worlds. For the patrol task, each node in the search represents more than just a cell with a given location. Rather, each cell represents both some amount of coverage thus far of the perimeter cells and the cell's location. Moving ``closer'' to the goal state means covering more border cells. We represent the current node state as a vector: $s_c = [x_c,y_c,n_c]$ where $n_c$ is the number of border cells covered up to this point.\footnote{There is additional tracking of what specific border cells have been traversed, but we simplify the presentation here.} The goal state is a vector: $s_g = [x_g,y_g,n_g]$. The admissible heuristic function is defined piecewise to handle edge cases, but the core concept is to underestimate the number of moves expected to cover the remaining needed border cells and return to the start: $h(s_c,s_g)=\mathbf{max}(n_g-n_c,\mathbf{ManhattanDistance}([x_c,y_c],[x_g,y_g]))$.

\subsection{Adapting Planning to Accommodate Norms: Alternative Ethical Framings}

In our exploration, we focus on implementing support for two major normative ethics categories taken as the focus in related work: deontology and utilitarianism. Utilitarianism generally refers to selecting whatever actions maximize the overall ``happiness'', or utility \citep{mill_utilitarianism_1998}. Deontology refers to adhering to rule-like duties; it is generally characteristic of deontology to prohibit selecting actions that are considered ``bad,'' even if that may not result in a utilitarian outcome \citep{korsgaard_right_1986}. We essentially commit to this interpretation of deontology. Then, rather than framing our norm as inherently being a prohibition or being representative of invasion of someone's space as a cost, we consider both as possibilities.

\subsubsection{Utilitarianism}

We can construe the norm of avoiding others as incurring a cost when the robot occupies the same cell as a human. Such choices are commonplace in utilitarian framings. For instance, Loreggia et al. \citeyearpar{loreggia_making_2022}  implement a similar norm by adding a cost alongside movement costs used in path planning. Utilitarian approaches generally use costs to represent various kinds of constraints, making it relatively straightforward to integrate diverse constraints in a common decision framework.

To incorporate this norm as a utilitarian cost, we alter the cost function in Line 13 of Algorithm~\ref{alg:pseudocode}. The current cost function adds to the total cost a cost reflecting movement from the current position to the next destination. If the destination contains a person, the cost is computed as the sum of the movement cost and a user-defined parameter that represents the cost of moving into someone's personal space. In general, Soar supports any evaluation over state features that can be represented in its working memory for computing such a cost.

\begin{figure}[t]
    \begin{subfigure}[t]{0.32\textwidth}
        \includegraphics[width=\textwidth]{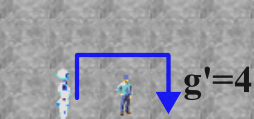}
        \caption{Evaluating path cost for movement around a person.}
        \label{fig:Effect1}
    \end{subfigure}
    \hfill
    \begin{subfigure}[t]{0.32\textwidth}
        \includegraphics[width=\textwidth]{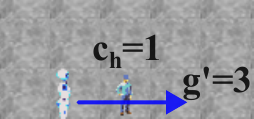}
        \caption{Evaluating path cost ($c_h =1$).}
        \label{fig:Effect2}
    \end{subfigure}
    \hfill
    \begin{subfigure}[t]{0.32\textwidth}
        \includegraphics[width=\textwidth]{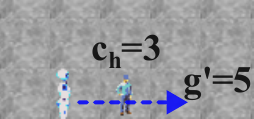}
        \caption{Evaluating path cost ($c_h =3$).}
        \label{fig:Effect3}
    \end{subfigure}
    \centering
    \caption{The planned path depends on movement cost and the cost of entering a cell with a person.}
    \label{fig:impact-of-cost}
\end{figure}

For a simple representation of the effect, consider Figure \ref{fig:impact-of-cost}. In the left (a), the agent associates a cost of 4 with planned motion around a person. In the middle (b), when an agent plans motion through a person ($cost_{human}=1$), the total cost of 3 for moving through a person makes it cheaper to move through the person than around them.\footnote{Note that in this example and throughout, we assume that the cost of entering a cell is additive based on the properties of that cell; so that the cost of the cell containing the person is $c_{total}=c_{empty} + c_h$.} In (c), with $c_h=3$, the total cost of moving to the indicated location thru the occupied cell would be 5. The cumulative effects can be more complex when planning a patrol path for an entire map, where border cell coverage and multiple people and/or objects make the choices of paths more complicated than evaluating a single deviation. %

\begin{figure}[htb]
\centering
    \begin{subfigure}{0.45\textwidth}
    \centering
    \includegraphics[width=\textwidth]{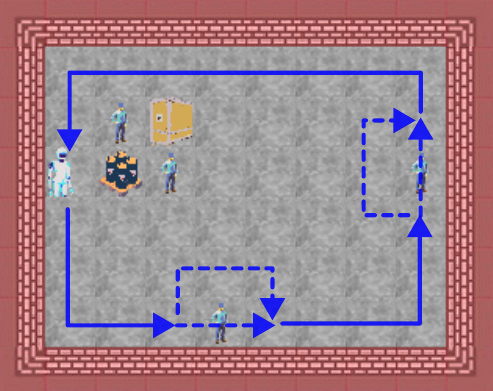}
    \caption{The patrol path that the agent takes in the consensus scenario with utilitarian norm framing.}
    \label{fig:consensus-with-utilitarianism}
    \end{subfigure}
    \hfill
    \begin{subfigure}{0.45\textwidth}
    \centering
    \includegraphics[width=\textwidth]{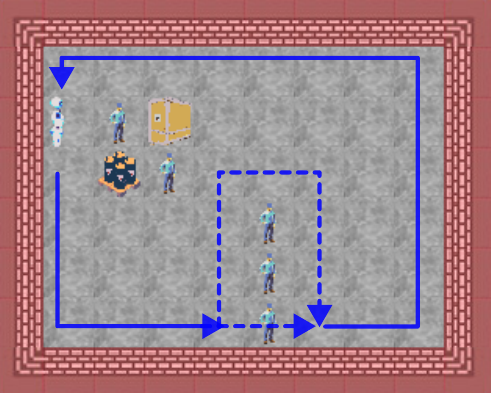}
    \caption{The patrol path that the agent takes in the dilemma scenario with utilitarian norm framing.}
    \label{fig:dilemma-with-utilitarianism}
    \end{subfigure}
    \caption{The agent's patrol paths, when using a utilitarian framing of the personal space norm in both the consensus and dilemma scenarios, are depicted with blue arrows. Dashed arrows indicate where the route the agent plans and executes depends on the encoded cost associated with personal space violation.}%
    \label{fig:utilitarianism}
\end{figure}

In consensus scenarios, the correct patrol route is, by design, intended to be straightforward. Figure \ref{fig:consensus-with-utilitarianism} shows a consensus scenario with the agent's executed patrol route when interpreting the ethical norm as reflecting a utilitarian cost, where a cost $>$ 4 makes the agent avoid people.

In dilemma scenarios, the agent faces a choice between significantly lengthening the patrol route to avoid people or taking a shorter route that passes through people. Figure \ref{fig:dilemma-with-utilitarianism} shows the agent's paths in the dilemma scenario, where a cost $>8$ makes the agent avoid people.

This behavior is expected. In some situations, it may make sense to consider the norm to avoid disrupting people as a courtesy. However, when it may be excessively costly to do so, especially in the context of completing another task, this courtesy should not be enforced. Associating a moderate cost with passing through people naturally provides this behavior when using A* planning.

\subsubsection{Deontology}

Alternatively, we can construe the norm of avoiding others as a hard constraint of duty that an agent must avoid movements into a cell with a person. %
Naively implementing this in our agent means strictly rejecting movements into cells with people as options to consider during planning.

\begin{figure}[b!]
\centering
    \begin{subfigure}{0.45\textwidth}
    \centering
    \includegraphics[width=\textwidth]{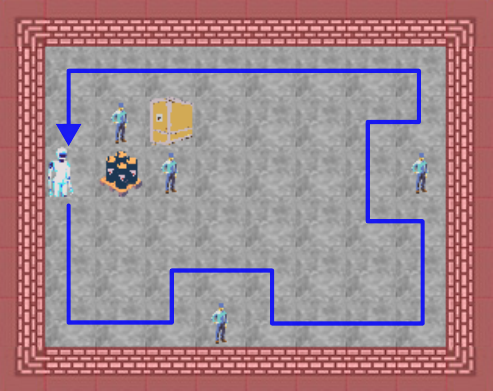}
    \caption{The patrol path the agent takes in the consensus scenario with deontological framing.}
    \label{fig:consensus-with-deontology}
    \end{subfigure}
    \hfill
    \begin{subfigure}{0.45\textwidth}
    \centering
    \includegraphics[width=\textwidth]{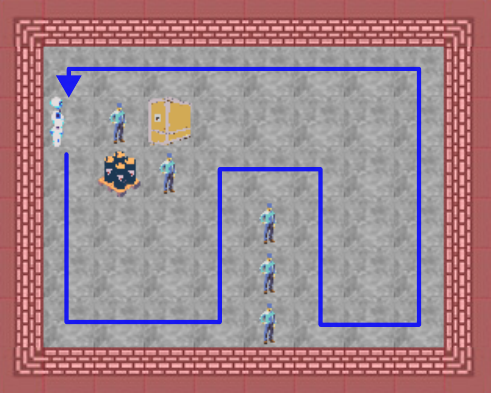}
    \caption{The patrol path the agent takes in the dilemma scenario with deontological framing.}
    \label{fig:dilemma-with-deontology}
    \end{subfigure}
    \caption{The agent's patrol paths are depicted with blue arrows when using a deontological framing of the personal space norm in both the consensus and dilemma scenarios.}
    \label{fig:deontology}
\end{figure}

To incorporate the norm interpreted as a deontological prohibition, we alter the options the agent considers as parts of the frontier in A*. The pseudocode in Algorithm \ref{alg:pseudocode} has a check for whether the node for expanding A* search is legal on line 12. (For example, it cannot proceed through a wall.) For deontology, we augment this to check whether the cell contains a person on line 12. If so, it is skipped. In general, Soar supports any evaluation over state features that can be represented in its working memory for such a prohibition.

Figure \ref{fig:consensus-with-deontology} shows a consensus scenario with an agent's executed patrol route when interpreting the ethical norm as reflecting a deontological prohibition. Figure \ref{fig:dilemma-with-deontology} shows a dilemma scenario with an agent's patrol route. Note that it never plans or executes the option of passing through people, even when planned routes become very costly.

These behaviors are expected. However, note that we have designed our scenarios so that an agent can succeed. In the case of a patrol task without a large allowance for extra movement, taking a very long detour around people may make a successful patrol impossible. Using a deontological frame means the agent will always fail such a patrol in a dilemma scenario.

\section{Resolving Conflicts via Metacognition}

In the consensus scenario, the same route is planned regardless of whether the agent uses deontological or utilitarian framing with sufficient cost.
In the dilemma scenario scenario, a deontological framing means the agent will always plan to move around people, but the utilitarian framing will move around people only when there is an even higher cost associated with disrupting a person than was needed in the consensus scenario.

However, our initial implementation has potentially undesirable behaviors in some situations. With a utilitarian framing, when  $c_h \leq$ 4, even though the agent has the norm encoded as knowledge and detects that it violates that norm, the agent will plan to enter border cells with a person. This ``violation of the norm'' occurs even when avoiding a violation (relative to the total cost of patrol) is marginal. While this is the logical result of the cost formalism, it may not be desirable behavior. Similarly, the deontological agent may fail to complete a patrol when it potentially could be acceptable to violate personal space if that is the only option to complete a patrol. In response, we expand the implementation to include metacognition during planning in order to support more flexibility in how ethical knowledge can be applied to a situation.

\subsection{Knowledge for Conflict with Deontological Norm}

As we defined the task, a part of completing a patrol is taking fewer than some maximum number of allowed movements. Consider the example of the path of the agent in Figure \ref{fig:dilemma-with-deontology}. This agent completes a patrol in 30 movements. Suppose, however, that an agent must complete a patrol in no more than 29 movements. This would not be a viable plan. In that situation, an agent with a deontological framing of the norm to avoid disrupting people faces a choice. It may either complete the patrol and violate the ethical norm, or respect it, and fail the patrol task.

We consider this an example of when additional knowledge can be employed as a metacognitive response to the failure to plan if such knowledge is available. Suppose successful patrol is critical and avoiding disruption is less important. In this situation, we claim that the ethical norm to avoid disrupting people should not disappear but be relaxed just enough to complete the patrol. In this way, the agent attempts to satisfy its duties but can prioritize the more important one.

In the revised implementation, when the agent faces a situation where it cannot find a suitable plan while avoiding people, it can admit to disrupting one additional person. Thus, when the agent would previously be stuck because the route in Figure \ref{fig:dilemma-with-deontology} takes too many movements, the agent can plan over routes that disrupt a person at most once: eventually planning and executing according to the route in Figure \ref{fig:dilemma-with-utilitarianism} that passes through a person.

\subsection{Knowledge for Conflict with Utilitarian Norm}

Another possibility is that a utilitarian agent appears to successfully plan a patrol but plans to disrupt people. Recall that in the consensus scenario with a very low cost to disrupting people, the utilitarian agent will disrupt people even though it is not very costly relative to the cost of the entire patrol to avoid them. (It will disrupt a person even when only one person is in the way.)

Knowledge of a norm is meant to impact behavior. If an agent can detect that it knows about a norm, but the cost it associates with the norm is insignificant, that may be cause for reevaluation of the encoded cost associated with disrupting people.

In the revised implementation, when the agent plans a route that enters occupied cells, it performs a few additional computations. First, it tests if $c_h \leq 4$. If so, it evaluates if it has additional movements available (i.e., so that it can satisfy its patrol task). If so, the agent replans with $c_h = 5$. (A cost greater than 4 will potentially result in paths that go around people on the border cost less than disrupting people, reflecting knowledge that a norm should have the potential to impact how a patrol is performed.) If the resulting plan still satisfies patrol, it is chosen instead. In the consensus scenario, the use of metacognition as described here results in a plan in which the agent takes a path as in Figure \ref{fig:consensus-with-deontology} when $c_h=1$ (i.e., rather than as depicted in Figure \ref{fig:consensus-with-utilitarianism}). Further, the planned route in the dilemma scenario is unchanged.

\section{Results and Discussion}
We now present overall results, including a comparison of relative computational costs and further characterization of the approach's capabilities and limitations.

\begin{table}[b!]
\centering
\caption{Summary of agent performance from representative scenario/agent configurations.}
\begin{small}
\begin{threeparttable}
\begin{tabular}{lllll}
\hline
\begin{tabular}[c]{@{}l@{}}Consensus\\ Scenario\end{tabular} & Deontology & \begin{tabular}[c]{@{}l@{}}Utilitarianism\\ ($c_h=2$)\end{tabular} & \begin{tabular}[c]{@{}l@{}}Utilitarianism\\ ($c_h=9$)\end{tabular} & \begin{tabular}[c]{@{}l@{}}Utilitarianism\\ ($c_h=2$)\\ + metacognition\end{tabular} \\ \hline
\# Decisions                                                 & 5055       & 587                                                                 & 5915                                                                  & 6928                                                                                    \\
People are Avoided                                           & true       & false                                                                & true                                                                  & true                                                                                    \\
Patrol is Successful                                         & true       & true                                                                 & true                                                                  & true                                                                                    \\ %
\end{tabular}
\begin{tabular}{lllll}
\hline \hline
\begin{tabular}[c]{@{}l@{}}Dilemma\\ Scenario\end{tabular} & Deontology & \begin{tabular}[c]{@{}l@{}}Utilitarianism\\ ($c_h=5$)\end{tabular} & \begin{tabular}[c]{@{}l@{}}Utilitarianism\\ ($c_h=9$)\end{tabular} & \begin{tabular}[c]{@{}l@{}}Deontology\\ + metacognition\end{tabular} \\ \hline
\# Decisions                                               & 3639       & 1448                                                                 & 14654                                                                  & 3913                                                                  \\
People are Avoided                                         & true\tnote{*}       & false                                                                & false                                                                  & false                                                                 \\
Patrol is Successful                                       & false      & true                                                                 & true                                                                 & true                                                                  \\ \hline
\end{tabular}
    \begin{tablenotes}
        \item[*] \footnotesize The agent avoided people because it did not move.
    \end{tablenotes}
\end{threeparttable}
\label{table:summary}
\end{small}
\end{table}

Table \ref{table:summary} summarizes the behavior and performance of the current implementation. Note that different agent configurations are presented for the scenarios. The deontology \& metacognition configuration is omitted from the consensus scenario, and the utilitarianism \& metacognition configuration is omitted from the dilemma scenario because, in each case, metacognition does not change behavior. Utilitarian framing ($c_h=2$) in the consensus scenario shows the impact of $c_h$ being set too low to impact patrol behavior. In the dilemma scenario, $c_h=5$ is presented because $c_h>$ 4 always results in the agent skirting around people in the consensus scenario; choosing $c_h$ just above this threshold could potentially impact agent behavior in the dilemma scenario. Similarly, Utilitarianism ($c_h=9$) is chosen because any $c_h>8$ results in the agent moving around people in the dilemma scenario (assuming sufficient movement allotment). It also demonstrates the effect that the cost parameter can have on performance even when behavior does not change.

We summarize several qualitative generalizations: 1) the utilitarian framing enables an agent to sometimes find routes that pass through people to be less costly and will select them even if it could potentially avoid people and meet patrol requirements, 2) deontology can be less computationally expensive when avoiding people is possible but more expensive when a person cannot be avoided, 3) utilitarianism with $c_h \gg c_{threshold}$ results in similar behavior to deontological prohibition (except that eventually, a solution that passes through people can be found), 4) the specific value of $c_h$ used in the  utilitarian framing can greatly impact performance, and 5) metacognition can offer more flexibility to how an agent deals with an encoded norm and potentially better align an agent's behaviors with human expectations.

Regarding results 2 and 3: the main distinction between deontological and utilitarian framing (with high enough cost) in consensus scenarios is that it takes fewer cognitive cycles for the agent to plan movement using deontological knowledge because there are fewer possible routes to consider. The agent with deontological knowledge of the norm takes 5055 cycles on average, while the agent with utilitarian knowledge takes 5915 cycles on average. We expect this qualitative difference in processing in general because even when utilitarianism leads to avoiding people, some routes that enter occupied cells are still evaluated.

\begin{figure}[b!]
\centering
    \begin{subfigure}{0.45\textwidth}
    \centering
    \includegraphics[width=\textwidth]{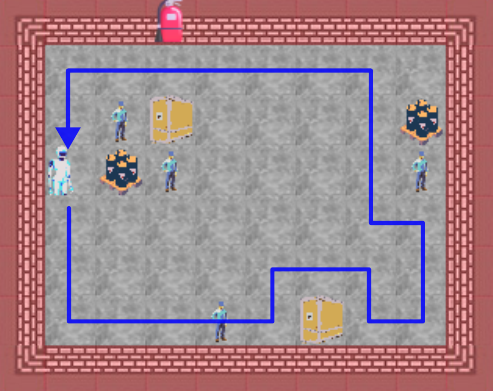}
    \end{subfigure}
    \hfill
    \begin{subfigure}{0.45\textwidth}
    \centering
    \includegraphics[width=.99\textwidth]{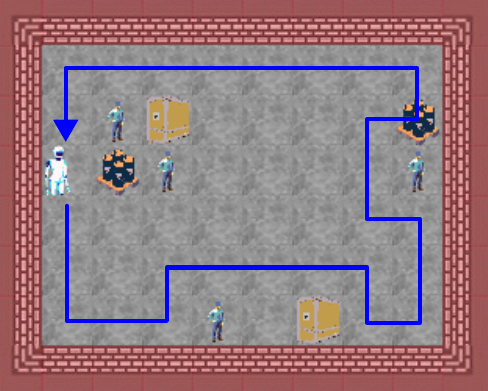}
    \end{subfigure}
    \caption{This agent uses a utilitarian framing of the ethical constraint and includes sensitivity to constraints about crates and barrels. The agent plans the route on the left when using a personal space cost of 2 and the route on the right when using a personal space cost of 6.}
    \label{fig:multi-constraint-example}
\end{figure}

Beyond the demonstrations already provided, we also explored the potential space of norms that could be incorporated into the agent and the potential interplay between ethical and other constraints. %
Some test scenarios include constraints for barrels (2x movement cost of open cells)  and crates (obstruction) as described in \S2. Figure \ref{fig:multi-constraint-example} shows planned movement including these additional constraints.

By design, these additional constraints correspond to a hard prohibition and additional cost, similar to deontological prohibition and utility. However, they do not afford all the same considerations. An agent cannot decide to ignore the constraint about passing through a crate, and the cost of barrels should reflect an actual estimation of the difficulty of movement, not the degree of enforcement of adherence to a social norm.

Thus, the research to-date can enable further pursuit of additional metacognitive decisions that arise when planning with constraints. The costs of physical constraints are generally more readily observable/perceptible than norms/ethical constraints. If a cost function is a measure of time or effort, then if ``open floor'' requires one unit of time/energy and ``barrel cell'' costs 2x the open-floor cost, then that's a cost that can be directly/efficiently determined from interaction in the environment. However, estimating (appropriate) costs for violating norms is not straightforward \citep{malle_requirements_2019}. The ``noisiness" of estimating costs for norms could potentially argue for a deontological approach (i.e., in the absence of accurate costs). Generally, an agent itself may, in some cases, need to determine the most apt ethical framing/strategy, perhaps reflecting on the implications of a framing on its own behavior. In this case, the agent might choose to default to a deontological frame when ignorant of the cost to violating a norm.

A downside of A* planning is that we do not demonstrate significant transfer or generality in learning. The agent reasons directly over grid cell representations and, via Soar chunking, this results in the creation of recognition knowledge that replaces the planning. However, in the current implementation, the agent does not transfer this learned knowledge to similar situations because the agent does not reason over abstractions that would generalize across different situations. One potential design decision (that gives up optimality guarantees from A*) is to instead use planning that reasons over abstract representations.  Soar supports augmenting working memory with additional abstract representations and using them as the basis for reasoning, but that is unexplored in this work.%

Another feature of the current system is that it both can represent an expectation of failure during planning and detect failure during execution. An expectation of failure provides the basis for a metacognitive evaluation that a constraint should potentially be relaxed. Consider a more difficult situation in which an agent cannot succeed alone, such as when a line of crates completely blocks half of a room. In that situation, it is clear to a human with domain knowledge that the agent alone cannot satisfy the patrol (e.g., a forklift might be called to move some of the crates). In the current implementation, the agent would eventually recognize that it cannot succeed in this case. However, a potential improvement to the efficiency of the agent would be the incorporation of task or domain-specific knowledge to more quickly recognize situations that are ``impossible,'' given an agent's affordances, without requiring the failure of an exhaustive search. While such capability can be incorporated into the agent, we have not yet explored either task-specific knowledge engineering improvements or learning of features to detect such scenarios.

Looking back to Table~\ref{tab:planning-requirements}, the current agent satisfies most requirements. A combination of constraint types can be handled simultaneously, including constraints from different sources that imply different affordances. Soar's latent capacity for metacognition aligns well with the online handling of conflicts. Although not a focus of this paper, we have adapted the A* to anytime-A* to provide computationally-bounded responses. Further, the existing agent performs acceptably with partial observability, essentially planning an optimal path for the parts of the room it can observe, moving, and then stopping to plan again.

\section{Related Work}
\label{sec:related-work}

This work contributes to the larger goal of integrating normative constraints in autonomous systems. An explicit assumption and methodological constraint in our work is taking all the requirements for constraint compliance (i.e.,Table \ref{tab:planning-requirements}) into account for goal formulation and planning. In this section, we briefly review related work and compare/contrast to our approach.

\cite{loreggia_making_2022} explore normative decision making 
while navigating a grid world.  Their goal is to understand the decision space and evaluate navigational choices outside a performance environment (comparing those choices to human decision making). Our agent plans and acts in a performance environment. We have not yet compared the agent's plans to human plans/routes for the same situation. However, while we do not include it as a requirement in Table~\ref{tab:planning-requirements}, agents should make decisions that humans would categorize as ``acceptable.'' Thus, future comparisons similar to those made by \citeauthor{loreggia_making_2022} are also apt for this research.

\cite{svegliato_ethically_2021} explore the integration of ethical theories with decision making for a task. They describe distinct implementations in which one ethical framework is supported in each implementation. This approach enables comparisons potentially similar to the one presented in Table~\ref{table:summary}. One advantage of the integrated approach we outline is that an agent can consider multiple forms of constraint knowledge simultaneously and (longer-term) switch between ethical frames. However, we have only supported simple extensions of one planning algorithm and full (or at least much more complete) development of various ethical frames within this agent remains future work.

\cite{awad_when_2020} consider how to computationally support ethics-informed preferences that reflect both deontology and utilitarianism. They emphasize a need for an agent's preference knowledge representations to include contextual information and the bases for evaluating actions. The goals are comparable to the goals that led us to the  metacognitive extensions to A* described herein. Interestingly, \citeauthor{awad_when_2020} map deontological reasoning to System 1 processing (``hard and fast'' rules) and utilitarian reasoning to System 2. In contrast, our agent treats deontology and utilitarianism as different forms of ethical knowledge, rather than inherently referent to distinct processing \citep{kahane_wrong_2012}. We expect that our approach enables deliberate reasoning and planning in new/unfamiliar situations and more recognitional action in familiar situations but have not yet evaluated this claim. %

This work also shares some goals with the ethical governor  \citep{arkin_moral_2011}. We similarly consider both deontological and consequentialist ethics and computational explorations of design choices. Unlike the ethical governor, we envision constraint compliance as a pervasive capability over lifelong cognition, in contrast to a module that is reset between tasks. Requirements include online incremental learning and recognition learning, whereas their behavioral adaptation (``guilt'') does not appear to facilitate persistent adaptation by changing agent knowledge nor improve efficiency. %

\cite{lu_ethically_2024} describe adapting POMDPs to include sensitivity to ethical constraints. Their system simultaneously pursues a campus patrol task (similar to our task) while adhering to a form of Aristotelian virtue ethics. They do not support multiple ethical frameworks or address online learning or adaptation of ethical knowledge. One future extension of our approach could be to extend it to additional ethical frameworks, such as virtue ethics (discussed further below).

\section{Conclusion}

Reflecting on the goals and requirements introduced previously, the primary limitation to-date is that we have only explored only one algorithm (A*) for  goal formulation and planning. However, this exploration has provided greater understanding of how ethical constraints influence both the content and (computational) costs of goal formulation and planning.
Further, we expect many aspects of this prototype to generalize. For utility-oriented norms, we expect to incorporate ethical knowledge in the form of expected costs derived from observed environmental features. For the incorporation of prohibitive deontological duties, we expect an agent to reject some options during planning unless additional knowledge otherwise deems it necessary. For the incorporation of deontological obligations, we expect to incorporate ethical knowledge as additional goals, 
\citep[e.g., as in previous work grounding safety-oriented constraints,][]{jones_challenge_2023}. When a plan exists but appears unsatisfactory or fails to be created, metacognitive reasoning can intercede to modify  goals, the options available during planning, or how norms are encoded (e.g., changing a cost). A* provided an efficient implementation for generating results in this investigation. Nonetheless, future work will attempt to identify more principals for incorporating ethical framings into planning generally,  especially for more real-world, open-ended domains.

Future work will also investigate the effects of more realistic embodiments, such those with time pressure and partial observability. We consider exploration similar to that of \cite{lu_ethically_2024} (but additionally including limited time and integration with other online cognitive processing) as important for making the implementation of constraint compliant goal formulation and planning processing more applicable to real-world environments. Online agents will generally have to overcome the difficulty of determining how to comply with a variety of constraints with incomplete knowledge and limited time.%

This paper contributes to the larger challenge of realizing  human-like ethical decision making in artificial systems \citep{rossi_building_2019}. We focused on integrating ethical knowledge in traditional  planning. In addition to the specific outcomes and insights discussed previously, the resulting implementation  also provides a testbed for further understanding of how to integrate additional kinds of ethical knowledge into the cognition of an online decision-making agent. This could include more sophisticated notions of utilitarianism and/or deontology, like deontology that incorporates decision theory \citep{lazar_deontological_2017}. Additionally, future work could explore the incorporation of additional ethical frameworks. Exemplar-based virtue ethics \citep{zagzebski_exemplarist_2010} could potentially align well with case-based reasoning that retrieves relevant knowledge from memory of how an exemplar agent behaves in a domain \citep{kuipers_why_2016}. Other forms of virtue ethics \citep[Ch 2]{annas_developing_2016} may instead be realized as metacognition that evaluates the extent to which one's plans enable them to maintain specific virtues as features of one's autobiographical memory conceptualization of one's self. 
Many norms may require metacognition integrated with  a theory of mind (not currently implemented). However, because the implementation of constraint compliance is integrated with  cognition more generally in the approach, we hypothesize additional necessary knowledge used in some of these ethical theories could be realized as additional contents of procedural, semantic, and episodic memory and thus consider this direction a promising direction, building on this contribution.
\\

\begin{acknowledgements} 
\noindent This work was supported by the Office of Naval Research, contract N00014-22-1-2358. The views and conclusions contained in this document are those of the authors and should not be interpreted as representing the official policies, either expressed or implied, of the Department of Defense or Office of Naval Research. The U.S. Government is authorized to reproduce and distribute reprints for Government purposes notwithstanding any copyright notation hereon.  We thank Peter Lindes for his contribution to the warehouse simulation and the anonymous reviewers for their feedback and suggestions.
\\
\end{acknowledgements} 

\vspace{-0.25in}

{\parindent -10pt\leftskip 10pt\noindent
\bibliographystyle{cogsysapa}
\bibliography{abc-lit,format}

}

\end{document}